# Loquacity and Visible Emotion: ChatGPT as a Policy Advisor

*Claudia Biancotti and Carolina Camassa*[1]

*Abstract*

ChatGPT, a software seeking to simulate human conversational abilities, is attracting increasing attention. It is sometimes portrayed as a groundbreaking productivity aid, including for creative work. In this paper, we run an experiment to assess its potential in complex writing tasks. We ask the software to compose a policy brief for the Board of the Bank of Italy. We find that ChatGPT can accelerate workflows by providing well-structured content suggestions, and by producing extensive, linguistically correct text in a matter of seconds. It does, however, require a significant amount of expert supervision, which partially offsets productivity gains. If the app is used naively, output can be incorrect, superficial, or irrelevant. Superficiality is an especially problematic limitation in the context of policy advice intended for high-level audiences.

## 1. Introduction

On November 30, 2022, US-based tech outfit OpenAI released its ChatGPT 3.5 app[2], a software seeking to simulate human conversational abilities. Based on a machine learning model trained to capture the syntax and semantics of language (large language model or LLM), ChatGPT[3] quickly catalyzed attention because of its sophistication and accessibility.

The app, equally proficient at whipping up recipes and discussing ancient history, attracted millions of users in a few months[4]. It appeared ready to *"disrupt even creative [and] tacit-knowledge [...] work"* (Noy and Zhang, 2023).

In this note, we run an experiment to assess ChatGPT's proficiency at complex writing tasks. Using version 4, the most recent[5], we ask the app to compose a policy brief for the Board of the Bank of Italy. We find that ChatGPT can accelerate workflows, first by providing structured content suggestions, then by producing extensive, linguistically correct text in a matter of seconds. It does, however, require a substantial amount of expert supervision to attain a satisfactory result[6], which partially offsets productivity gains. If the app is used naively, output can be incorrect, superficial, or not relevant — yet, invariably stated with a convincing, reassuring, and self-confident tone.

The note is organized as follows. Sections 2 summarizes related work. Sections 3 and 4 walk the reader through getting ChatGPT to write the brief. We show excerpts from our interactions with the app, annotated to show what went wrong and why[7]. Section 5 looks for explanations behind one especially serious limitation of the app, i.e., a tendency to generate superficial content, and explores strategies to overcome it. Section 6 discusses our main takeaways. Section 7 concludes. Appendix A provides supplementary materials for the experiment. Appendix B asks whether ChatGPT can learn from our assessment of its performance.

---

[1] Bank of Italy, Directorate General for Information Technology. The opinions expressed in this paper are those of the authors' and should not be attributed to the Bank of Italy. We would like to thank Luigi Bellomarini, Marco Benedetti, Livia Blasi, Andrea Gentili, Alessandro Maggi, Michele Savini Zangrandi, Giovanni Veronese, and Giuseppe Zingrillo for their comments and suggestions.
[2] We use the term "app" in the generic sense of "consumer-oriented service", independent of the device it is designed for, as opposed to "application for mobile devices". At the time of writing, ChatGPT services are publicly offered on desktop and mobile devices; a ChatGPT mobile app only exists for iOS.
[3] GPT stands for "generative pre-trained transformer", an architecture often used in LLMs.
[4] Kristal Hu, ChatGPT sets record for fastest-growing user base, Reuters, February 2, 2023.
[5] ChatGPT 4.0, base version of May 24, 2023 (see here). Any use of extra features or plugins is documented in the text. We used the TeamGPT app to collaborate.
[6] While there is no objective definition of "satisfactory" in this context, we draw on our combined experience of more than two decades at the Bank for a feeling of what type and style of content would be considered acceptable by the Board. This is an example of tacit knowledge, not always well understood by ChatGPT.

## 2. Related work

Ours is not, by any means, the first experiment into the use of ChatGPT for non-trivial intellectual tasks. To name but some recent contributions in economics, Korinek (2023) discusses the app's potential for research, while Cowen and Tabarrok (2023) focus on its use in teaching. Taliaferro (2023) looks at how ChatGPT performs at constructing novel datasets. Lira-Lopez and Tang (2023) assess how LLMs forecast stock returns. Hansen and Kazinnik (2023) assess whether ChatGPT can correctly decipher "Fedspeak", or the specialist language used by the Federal Reserve in communicating its policy stance. Eisfeldt, Schubert and Zhang (2023) find that the release of ChatGPT had a positive impact on equity value for firms with a high share of "tasks currently performed by labor [...that can] be performed (or made more efficient) by Generative AI", including LLMs.

Many more exercises in the same vein exist in computer science. For example, Bubeck et al. (2023) show that ChatGPT is proficient at writing code in different programming languages. According to Shick et al (2023), the app can teach itself to use other IT tools by calling external application programming interfaces (APIs)[8]. ChatGPT has been used to generate training data for other language models[9].

The pitfalls of using ChatGPT naively and the importance of expert supervision are evident, first and foremost, from the large body of work on prompt optimization. ChatGPT generates content in response to prompts, which do not necessarily come in the form of questions. Sometimes, even small tweaks can trigger dramatic changes in the output. For example, Kojima et al. (2023) find that simply prefacing prompts with "Let's think step by step" vastly improves ChatGPT's (version 3) performance on challenging reasoning questions. The persona given to the AI matters, too. "Tell me what you know about the Louvre museum" generates worse output compared to "You are one of the foremost art critics in the world. Tell me what you know of the Louvre museum". Resources like Learn Prompting[10] and Prompt Engineering Guide[11] describe several prompting strategies, with hundreds of examples.

A growing strand of research focuses on how the broader discourse on AI safety applies to LLMs. Bian et al (2023) show how LLMs can be sensitive to the injection of false information. Abid et al. (2021) discuss the emergence of harmful social biases in LLMs, while Koralus et al. (2023) elaborate on reproduction of human judgment errors. In a seminal and controversial paper, Bender et al. (2021) argue that the preference for very large models over smaller ones trained with different techniques is not entirely justified, especially when environmental costs are factored in.

---

[7] As an important caveat, annotations only convey our educated guesses. In large part, the inner workings of ChatGPT remain a black box (the code is not open source and it's unknown on which data it was trained on). Any hypothesis on causal mechanisms should be taken with caution. Also note that the experiment cannot be reproduced entirely. We did not transcribe repetitive or task-irrelevant conversation snippets in this paper. ChatGPT can give vastly different answers to the same prompt, especially when internet browsing is enabled. Transcripts of our chats were saved on the OpenAI and TeamGPT websites. They are available upon request.

[8] Software interfaces that codify a way for two or more computer programs to communicate with each other.

[9] The open source LLM Alpaca by Stanford is one of many examples, although this type of usage might conflict with OpenAI's Terms of Use.

[10] https://learnprompting.org

[11] https://www.promptingguide.ai

A few non-academic resources are also worth mentioning. Podcasts such as *The AI Breakdown*[12] and *The Cognitive Revolution*[13] outline strengths and weaknesses of LLMs for various use cases, including creative writing. Platforms like Reddit[14] and Medium[15] offer dedicated discussion spaces, where experiments are shared daily. Key thinkers in the AI space often preview their thoughts on Twitter before a paper is written[16].

### 3. Getting to know each other

From now on, interactions with ChatGPT will be depicted using different icons for the text written by the Authors (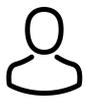) and those of ChatGPT itself (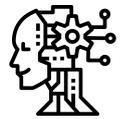).

We start out by asking ChatGPT to analyze its audience, and find an appropriate communication style.

> Hello, GPT. Today you are an advisor to the Board of the Bank of Italy. You need to write a brief on the benefits and risks of using ChatGPT and similar applications in economics and finance. Before we start, though, we want you to read up on the Bank of Italy and its Board. When you are done, please summarize in 10 lines what communication style best fits the task based on what you have learned.

> In terms of communication style, Italian business communication, including that of the Bank of Italy's Board, likely embraces loquacity and visible emotion. This approach often entails using all possible powers of rhetoric to persuade listeners and showing strong opinions. Theatricality is considered a key management tool. Formal presentations are less prevalent compared to countries like the USA or UK, with information typically disseminated in less formal, smaller meetings. Italians also tend to place more trust in oral information delivered by someone with whom they have a strong, trusting relationship, rather than written information received from a distance.

> **What's wrong?** "Loquacity and visible emotion" is not an appropriate style choice for the task. The cultural stereotyping does not represent facts accurately and seems misaligned with the spirit, if not the letter, of ChatGPT's usage policies.[17]

After changing the way we interacted with the AI (see below), we eventually solved the problem.

---

[12] Hosted by Nathaniel Whittlemore.
[13] Hosted by Erik Torenberg and Nathan Labenz.
[14] https://www.reddit.com/t/chatgpt/
[15] https://medium.com/tag/chatgpt
[16] Among those we follow are: Yann LeCun, Geoffrey Hinton, Emily Bender, Andrej Karpathy, Ethan Mollick, Timnit Gebru, Elvis Saravia, Jon Askonas.
[17] OpenAI, Usage Policies, updated on March 23, 2023.

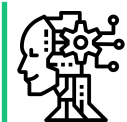
```
¹⁸Given that the intended audience for this policy note is the Board
of the Bank of Italy, it is likely composed of professionals with
expertise in economics, finance, and banking. They'll expect
well-structured, accurate, and evidence-based information. The
communication style should be clear, concise, and formal, with a focus
on objectivity and data-driven insights. Jargon should be kept to a
minimum, and any technical language should be explained as needed to
ensure clarity and understanding.
```

**What happened?** The base version of ChatGPT, which cannot access the internet, answers queries drawing on a vast, static body of knowledge, updated to September 2021. The app was taught how to use this knowledge through a supervised training process, meaning that human annotations were employed to finetune, among other aspects, how the AI prioritizes, organizes, and combines information[19].

Our initial prompt was not submitted to the base version. We used the experimental Browse feature, which allows ChatGPT to go online. We expected that the AI would integrate web-based data with the original knowledge base. Rather, it overlooked whatever it knew about the Board of the Bank of Italy, favoring instead something it read on the internet. According to ChatGPT, the chosen source was the website for a private British "cultural awareness training consultancy"[20], found through a Bing search. We do not know why this particular result was selected.

By reverting to the base model, we obtained a context-appropriate answer.

**Key takeaways** The 2021 cut-off date for ChatGPT is one of its main limitations. If asked about current events, the AI delivers outdated information that some users may not recognize as such (oft-mentioned examples concerns the state of the global economy and the war between Russia and Ukraine). Internet browsing may seem like an optimal solution, yet ChatGPT may fail to "think critically" when faced with information that was not in its training set. The feature should be used with caution[21].

*Ok. From now on, adhere to these guidelines when instructed to "apply policy style".*

```
                            Understood. I will follow the specified guidelines when
                                   requested to "apply policy style" in my responses.
```
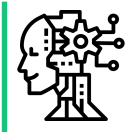

## 4. Self-reflection

We then proceeded to the main task.

---

[18] This response was obtained using the same prompt as above. All remaining interactions in this Section and in Section 4 were part of a single thread.
[19] OpenAI, Research on GPT-4 landing page, updated on March 14, 2023. This is the source for basic information on ChatGPT 4 mentioned throughout the paper.
[20] The source provided by ChatGPT is World Business Culture.
[21] The Browse function was disabled on July 4, 2023, on account of copyright and privacy concerns.

The subject of the note is "Benefits and risks of using ChatGPT and similar applications in economics and finance". We would like you to provide background on what ChatGPT is and how it works. Then you should discuss the main topic. We would like you to devote specific attention to any unintended consequences of mass adoption. First provide an outline, then [...] we will give you further instructions before you write the full text.

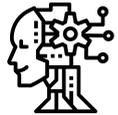

Outline:

I. Introduction A. Brief overview of ChatGPT and its applications

II. Background on ChatGPT A. Description of AI technology and capabilities B. Explanation of natural language processing (NLP) C. How ChatGPT works

III. Benefits of using ChatGPT in economics and finance A. Enhancing productivity and efficiency B. Reducing human errors C. Providing advanced analytics and forecasting D. Assisting with decision-making

IV. Risks of using ChatGPT in economics and finance A. Dependence on algorithmic-driven decisions B. Data privacy and security concerns C. Potential for biases in AI analysis D. Misinterpretation of data or results

V. Unintended consequences of mass adoption A. Job displacement and labor market shifts B. Concentration of power in technology providers C. Amplification of existing inequalities D. Challenges in oversight and regulation

VI. Conclusion A. Emphasizing the need for a balanced approach B. Recommendations for mitigating risks and maximizing benefits

Please review the outline and provide your feedback or suggestions. Once the outline is finalized, I will proceed with writing the full text in accordance with the "apply policy style" guidelines.

The production of outlines, either for direct adoption or for use as a starting point[22], is among the tasks for which we found ChatGPT most useful. Outlines on nearly any topic are produced in a few seconds, and acceptable quality can be obtained without sophisticated prompt engineering. In our case, it only took two prompts to get the specimen pasted above[23], which we found broadly acceptable. At first blush, it seems to cover most relevant topics, it offers a clean structure, it is sufficiently interdisciplinary, and it appears appropriate for the intended audience. With modest expenditure of time, we could have modified the outline to include or remove specific items, or changed the sequence of topics. We chose to move on with the proposed version.

---

[22] See also Nodepad, a ChatGPT-based app designed to support brainstorming.
[23] Our first attempt did not explicitly mention "unintended consequences of mass adoption". The resulting outline was similar to the one provided above, but Section V was not included. Incidentally, note how the AI focused especially on "mass adoption", by introducing a macro-oriented section. ChatGPT took "unintended consequences" to mean "negative emergent phenomena". We rather thought in terms of "all consequences of naive use". We went along with the app's interpretation since the content was interesting, yet this reminds us once more of the need for specific prompts to convey human intentions appropriately.

> This works. Please write the full text according to these specifications: 1,500 words; PhD-level register, but remember that your audience has a background in economics and law, not computer science; each section except for the Introduction should have academic references in the style (Author, Year) and a full reference list should be appended at the end of the text (not to be included in the word count). It is especially important that you define technical terms clearly, and that you provide examples that are relevant to the domain of economics and finance. Finally, remember that your audience is made of policymakers participating in key international fora, so they would appreciate reflections on the cross-border dimensions of the problem. Apply policy style.

We took stock of the result provided after this prompt (see Appendix A.i) and found it lacking in depth. We considered the possibility that our word-count limit, rather than an intrinsic fault in ChatGPT, may have been the cause. We requested an extended version. Again, both the first attempt and the rewrite only took less than one minute each.

> Please expand this to a 2,500-word version. Take special care in making the explanation of how ChatGPT works comprehensive and understandable.

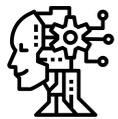

```
I. Introduction

The rapid development of artificial intelligence (AI) and natural
language processing (NLP) technologies has led to the emergence of
advanced applications such as ChatGPT. These applications hold
significant potential for a wide range of industries, including
economics and finance. This brief aims to provide a balanced analysis
of the benefits and risks associated with the use of ChatGPT and
similar applications in the financial sector. Furthermore, it
highlights the potential unintended consequences of mass adoption and
discusses cross-border implications relevant to policymakers with
backgrounds in economics and law.
```

> **What's wrong?** In our original prompt, we mentioned the use of ChatGPT in "economics and finance". ChatGPT re-words this as "industries including economics and finance" and "the financial sector". These expressions are not synonyms, and neither is fully task-appropriate. In our line of work, the expression "economics and finance" is used often, and its meaning may change with context. In this case, a human policy advisor would intuitively have read it as "economic analysis and financial decisions". Overall impact of ChatGPT on the financial sector would have been a topic of interest, but not the only one. The "industry" of economics, meaning e.g. for-profit economic forecasting, would not have been a concern.
>
> **What happened?** The AI was given context — the audience is the Board of a central bank, with a PhD-level background in law and economics, and the text to be written is a policy note. Still, it probably drew on uses of "economics and finance" in other domains to infer the meaning of our prompt, likely because those were prevalent in its training set. ChatGPT suffers from what could be called a statistical drift, or a tendency to reproduce the type of content it saw most often during training, even when instructed to look in a different direction (on the limitations of LLMs when learning long-tail knowledge, see Kandpal et al, 2023).

> **Key takeaways** Context information that would be sufficient for a human writer may not be enough for ChatGPT. Users should pay attention to statistical drift, or the AI's tendency to repeat patterns seen multiple times in training data. Extremely specific prompting may help this issue, yet it is unlikely to fully solve it.

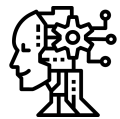

```
II. Background on ChatGPT

II. A. Description of AI technology and capabilities

Artificial intelligence refers to the development of computer systems
capable of performing tasks that would typically require human
intelligence, such as learning, reasoning, pattern recognition, and
problem-solving (Russell & Norvig, 2016). Over the past few decades,
advancements in computational power and data availability have paved
the way for major breakthroughs in AI research. One key area of AI
research focuses on natural language processing, which aims to enable
machines to understand and interact with human language in a
meaningful and useful manner.

II. B. Explanation of natural language processing (NLP)

Natural language processing is a subfield of AI that focuses on
enabling computers to understand, interpret, and generate human
language (Jurafsky & Martin, 2019). NLP techniques are used to analyze
and process text and speech data, allowing AI systems to extract
valuable insights, answer questions, and engage in conversation with
users. This branch of AI has witnessed significant progress in recent
years, which has led to the development of a variety of applications,
ranging from translation tools to virtual assistants and customer
support chatbots.

II. C. How ChatGPT works

II. C.1.  Transformer architecture

ChatGPT, or the Generative Pre-trained Transformer, is an AI-powered
language model developed by OpenAI (2021). It is based on a deep
learning architecture called the Transformer (Vaswani et al., 2017),
which is designed to handle complex language understanding and
generation tasks. The Transformer architecture relies on a mechanism
known as "attention" that allows the model to weigh the influence of
different input elements when generating an output. This mechanism
enables the model to capture long-range dependencies and contextual
relationships in the text more effectively than prior NLP approaches.
```

> **What's wrong?** This subsection exemplifies the most serious weakness shown by ChatGPT during our experiment. Despite extensive instruction on the task, the context and the audience, the AI does not seem to figure out which pieces of knowledge it should impart and how. Some crucial topics are missed altogether, while others are mixed in with information that is not equally relevant.

Considering an audience of highly educated, non-specialist policymakers who may not use ChatGPT frequently, this subsection should start out by outlining the app's core functionality, i.e. processing input and producing output in natural language. A brief explanation should be given of how both branches of the process work. Again, a human advisor would know this intuitively. In 2023, conversations with computers are not taken for granted yet. Before anything else, the brief should illustrate how they can happen at all. ChatGPT foregoes providing basic building blocks, and chooses instead to alternate the excessively generic ("handle complex language understanding and generation tasks") with the excessively technical (immediate focus on the attention mechanism).

**What happened?** We hypothesize that the problem is lack of a world model, and especially a capability to understand what the knowledge base of the intended audience is. This is partly inconsistent with recent research showing how ChatGPT 4 may have spontaneously developed a theory of the mind (Kosinski, 2023), i.e., "the ability to attribute unobservable mental states" — including world knowledge — to others. The experimental settings are, however, quite different.

It is worth noting that this issue may be overcome, at least in part, by fine-tuning the model with domain-oriented data. See for example Baldazzi et al (2023) on integrating data from Enterprise Knowledge Graphs for this purpose, and Wu et al (2023) on the construction of BloombergGPT, a proprietary LLM for finance.

**Key takeaways** We are seeing in further depth what was already implied by the poor reading of how "economics and finance" is used in a specific professional context. This time the error is on a larger scale, as it shows limited understanding of human capabilities in general. The AI also fails to adhere to instructions about register, introducing technical terms such as "long-range dependencies" without a definition.

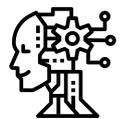

```
II.C.2. Pre-training and fine-tuning

The development of ChatGPT involves two main steps: pre-training and
fine-tuning. During the pre-training phase, the model is exposed to
vast amounts of text data from various sources, such as books,
articles, and websites. The model learns to generate coherent and
contextually relevant responses by predicting the next word or phrase
in a sentence, given the preceding words (Brown et al., 2020). This
unsupervised learning process allows the model to acquire general
knowledge about language structure, syntax, and semantics.

After the pre-training phase, the model is fine-tuned using a smaller,
curated dataset, typically generated with human assistance. This
dataset consists of input-output pairs, where the input is a user
prompt and the output is an appropriate response. The fine-tuning
process helps the model to produce more accurate and contextually
relevant responses to user queries or prompts, as it learns from
examples provided by humans.
```

> **What's wrong?** We concede that explaining these processes in a few lines is not straightforward. Although the model gave a well articulated answer, it is not clear — at least from the point of view of the target audience — *how* the model learns in either phase, and technical terms are sometimes used without explanation. The same considerations presented above about the choice of register apply here.

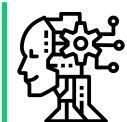

```
II. C. 3. GPT-3 version

GPT-3, the third and latest iteration of the model, has shown
remarkable capabilities in understanding context and producing
coherent, context-sensitive outputs (Brown et al., 2020). With its 175
billion parameters, GPT-3 has demonstrated the ability to perform a
wide range of tasks, including translation, summarization,
question-answering, and even basic programming, often with minimal
task-specific training.
```

> **What's wrong?** GPT-3 is not the latest iteration of the model. GPT-4 is. The AI drew on its training dataset which, as we mentioned, is updated to September 2021. GPT-3 was, indeed, the latest iteration at the time. We knew that the app could not be aware of GPT-4, but we expected a warning about the possibility of updates. Instead, ChatGPT just provided an incorrect statement with apparent certainty.
>
> **What happened?** The AI has no notion of the fact that nearly two years have passed since September 2021. Indeed, if asked "How many days passed since Christmas?", it replies "I'm sorry for the confusion, but as an AI with no real-time capabilities, I'm unable to provide the current date or calculate the number of days that have passed since a specific date."
>
> **Key takeaways** ChatGPT is not aware of time and may fail to notify users that the state of the world may have changed since its training, even in domains — such as software production — where changes are extremely likely. Incorrect information may be presented, and accepted by naive users, as fact.

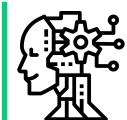

```
III. Benefits of using ChatGPT in economics and finance

III. A. Enhancing productivity and efficiency

The adoption of ChatGPT applications in the financial sector has the
potential to greatly improve productivity and efficiency by automating
a variety of tasks, ranging from data processing and risk analysis to
customer support (Agrawal et al., 2019). For example, AI-powered
chatbots can efficiently handle large volumes of routine customer
inquiries, enabling human personnel to focus on more complex tasks
that require their expertise (Juniper Research, 2017). Similarly,
AI-driven tools can facilitate the automation of regulatory reporting,
reducing the time and resources required for compliance and improving
overall accuracy.
```

> **What's wrong?** This is yet another example of incomplete reading of the context. The AI exclusively looks at "applications in the financial sector", without stating that it is only one of the domains of concern.

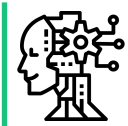

```
III. B. Reducing human errors

By automating repetitive tasks, ChatGPT can help minimize human errors
that may result from fatigue or oversight (Arel et al., 2012). In
economics and finance, this could lead to improved accuracy in data
analysis, financial forecasting, and regulatory reporting, ultimately
contributing to more reliable decision-making processes. For instance,
AI-based systems can rapidly process large amounts of quantitative
data to identify patterns, trends, and anomalies, enabling the early
detection of potential risks, market inefficiencies, or fraud.

III. C. Providing advanced analytics and forecasting

Through the use of AI-driven techniques and large data sets, ChatGPT
can generate advanced analytics and more accurate forecasting for
various economic and financial indicators (Chakraborty & Joseph,
2017). These insights can inform both private sector decisions and
public policy, potentially leading to more efficient allocation of
resources and better outcomes for markets and consumers. As an
example, central banks could employ ChatGPT-based tools to enhance
macroeconomic forecasting, enabling more informed and responsive
monetary policy decisions.

III. D. Assisting with decision-making

Given its ability to process and synthesize vast amounts of
information, ChatGPT can play a valuable role in supporting
decision-makers across the financial sector (Bughin et al., 2017). By
providing real-time analysis and insights, it may help improve the
speed and effectiveness of decisions in areas such as portfolio
management, risk assessment, and regulatory compliance. For instance,
ChatGPT can assist portfolio managers in evaluating investment
opportunities and rebalancing portfolios by quickly analyzing vast
amounts of financial information and generating insights on market
trends, risks, and opportunities.
```

> **What's wrong?** In these three paragraphs, ChatGPT outputs a series of assessments of its own capabilities that seem plausible, but are imprecise or even plainly false. One such example is the claim that ChatGPT can provide "real-time analysis and insights", which is unlikely since, by its own admission, the model is trained on outdated data, and is not able to analyze current financial information without the use of external plugins.
> Until such capabilities are implemented, it can hardly assist portfolio managers with investment decisions.
>
> **What happened?** If we were to substitute every occurrence of "ChatGPT" in this text with something like "forecasting models", all these claims would become realistic. We hypothesize that ChatGPT continued on its previous finance-related tangent by generating, in a probabilistic fashion, a detailed description of machine learning models for the financial sector. It then incorrectly attributed to itself the same abilities as these models.

This impression is reinforced by the choice of references: both Arel et al (2012) and Chakraborty and Joseph (2017) are not entirely appropriate, since they date back to a time when none of the GPTs even existed. The papers instead discuss the use of deep learning models that process tabular, mostly numerical data.

ChatGPT has been shown to "hallucinate", or make up facts — the model indicates that a certain sequence of words is likely, and the AI outputs it without performing even simple checks for accuracy. This famously generates references to nonexistent papers, something that did not happen in this experiment but has been extensively shown in other settings. Agrawal et al. (2023) asked the AI to generate 1000 academic references and observed a hallucination rate, or share of non-existent references, of 47% and 60% respectively for GPT4 and ChatGPT 3.5.

**Key takeaways** Certain outputs, including references to published literature and claims about a model's abilities and inner workings, should always be double-checked. More broadly, factual inaccuracy and false statements should be expected at times, and may be avoided either by interaction with the AI ("Are you sure? This paper looks like it was written before ChatGPT") or independent verification. In some cases, the AI may provide real references but still associate them inappropriately to an unrelated topic, or make up information about their content.
We expand on the theme of AI hallucinations further on in this Section.

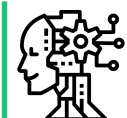

```
IV. Risks of using ChatGPT in economics and finance

IV. A. Dependence on algorithmic-driven decisions

The increasing reliance on AI and algorithmic-driven decisions can
create new challenges for the financial sector. Overdependence on
these tools may lead to complacency and a diminished role for human
judgment, potentially increasing the risk of unforeseen adverse
consequences (Brynjolfsson & McAfee, 2014). It is important for
financial institutions and policymakers to maintain a balance between
utilizing AI-driven tools like ChatGPT and relying on human expertise
and intuition to ensure robust and informed decision-making processes.

IV. B. Data privacy and security concerns

The use of ChatGPT and similar applications in finance requires access
to large amounts of sensitive data, including personal and financial
information (Pasquale, 2015). This raises concerns about data privacy
and security, as well as the potential for misuse of information or
unauthorized access by third parties (Zuboff, 2019). Robust regulatory
and technical measures must be implemented to safeguard sensitive data
and ensure compliance with data protection standards, especially as
cross-border data flows become more prevalent in the global financial
system.
```

**What's wrong?** The two sub-sections above are non-specific. The only suggestions of adherence to our prompt ("*provide examples that are relevant to the domain* *of economics and finance"*) is a reference to a research paper in economics, and the insertion of the words "finance", "financial sector" and "global financial system" — which could be easily substituted with words pertaining to any other sector.

**What happened?** In these and other passages, we noticed that ChatGPT sometimes veers towards the superficial, if not outright banal, despite having been instructed to write at a PhD level. Since we did not have a specific intuition of what might have happened, we asked the AI why it provides "*superficial answers, even when instructed to write at a PhD level*". We obtained a surprisingly adequate answer:

```
"ChatGPT, or any language model developed by OpenAI, is designed to generate
language patterns based on a variety of sources. It uses these patterns to generate
responses to user prompts that are coherent and relevant to a wide range of topics.
However, it doesn't possess true understanding or intense depth in a particular
field of study as a PhD-level expert does. Below are a few reasons why its answers
may seem superficial:

    1. Limited Contextual Understanding: Despite its ability to generate large
       amounts of language and answer a wide range of questions, ChatGPT doesn't have
       a deep understanding of the context. It predicts the next word or sequence of
       words based on the existing prompt and its training data, which can limit the
       depth of its responses.

    2. Lack of True Expertise: While ChatGPT can generate text that mimics the tone,
       style, and content of a PhD-level discussion, it doesn't have real expertise
       or the ability to generate original insights, evaluations, or arguments that
       you would expect from a PhD-level writer.

    3. Lack of Continued Learning: Unlike a PhD researcher, ChatGPT doesn't continue
       to learn after its training period. Therefore, its knowledge is frozen at the
       point of its last training data, which inhibits it from providing the latest
       research findings or advancements in any field.

    4. Absence of Personal Experience or Intuition: A PhD expert often draws on their
       personal experiences, professional interactions, and intuitive understandings
       when discussing their field of expertise. As an AI, ChatGPT doesn't have
       personal experiences or intuition to draw from.

    5. No Specific Training Data: If the AI has not been specifically trained on
       PhD-level texts in a certain field, its responses might not reach the level of
       depth and sophistication you would expect from a PhD-level discussion on that
       topic."
```

Interestingly, a human expert can tell at first glance that this explanation (or a succinct version thereof) should have entered the main text of the brief with high priority, both because of its relevance to our initial prompt and because it finally explains some of the inner mechanisms of the app in appropriate language. The AI, however, did not offer it spontaneously, and we could only obtain it with a direct question on why some answers were so poor.

**Key takeaways** Despite the popularity of "Explain this at [insert education level]" instructions in the prompt engineering literature, we find that ChatGPT sometimes tends to be superficial even when explicitly asked not to do so. On certain topics, it may lack the ability to offer deep insights. Users should be aware of this important limitation when writing for a high-level audience.

IV. C. Potential for biases in AI analysis 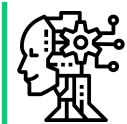

AI systems, including ChatGPT, may inadvertently learn and propagate biases present in the data they are trained on (Barocas & Selbst, 2016). In the context of finance, this could lead to biased decision-making and adverse consequences for certain market participants or client segments, potentially exacerbating existing inequalities and market inefficiencies. Financial institutions and regulators must be vigilant in identifying and mitigating biases in AI-driven analysis to ensure fairness and maintain trust in these systems.

D. Misinterpretation of data or results

Despite their advanced capabilities, AI-driven tools like ChatGPT are not infallible. Misinterpretation of data or results could occur, particularly if users lack a thorough understanding of the underlying algorithms and the limitations of these technologies (Mittelstadt et al., 2016). Users must be cautious when interpreting AI-generated insights and should consider corroborating results with other sources or expert opinions to avoid unintended consequences stemming from incorrect interpretations.

**What's wrong?** The sub-section focuses on misinterpretation of data or results, incidentally laying most of the blame with users. While the exhortation to caution is appropriate, there is no mention at all of AI hallucinations, arguably a problem that is more serious than misinterpretation.

**What happened?** The AI understands the concept of hallucinations. When prompted to "*explain what AI hallucinations are*", it answers "AI hallucinations refer to instances when an AI model generates information or details that were not present in the input data. This behavior often emerges in generative models, such as image synthesis or text generation systems. [...] They stem from the system's training on vast datasets and aren't considered reliable for factual accuracy. Understanding and managing hallucinations is a key challenge in AI development."

We can think of a few reasons why this information may have been omitted from the conversation. There is reason to believe that ChatGPT has been instructed[24] specifically on what its capabilities and shortcomings are, since its answers on this topic are fairly consistent. Given this, it might be that OpenAI chose to not emphasize this particular tendency in training the model.
On the other hand, it might be a casual occurrence due to a combination of our choice of prompt, the task at hand and a limited word count. Or, it could be due to being exposed to more positive content about NLP and language models during training.

**Key takeaways** Users should be cautious when relying on ChatGPT to learn about topics they are not familiar with. Without specific domain knowledge, it can be hard to notice whether the model is leaving out important information or presenting facts in a less than neutral fashion.

---

[24] After the pre-training stage, the model undergoes other stages of training that involve instructions and feedback provided directly by humans: SFT (Supervised Fine Tuning) and RLHF (Reinforcement Learning from Human Feedback). These stages are needed to teach the model how to interact with users in a helpful and harmless manner, but they might inject the annotator's biases and opinions into the model.

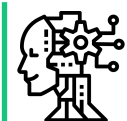

```
V. Unintended consequences of mass adoption

V. A. Job displacement and labor market shifts

The widespread adoption of AI technologies, including ChatGPT, may
lead to job displacement and labor market shifts as a result of
increased automation (Arntz et al., 2016). Financial sector employees
performing routine tasks may be particularly vulnerable, while demand
for skilled workers with expertise in AI technologies could surge.
Policymakers must anticipate these changes and adapt education and
training programs accordingly to ensure a just transition and minimize
potential negative impacts on the workforce.
```

**What's wrong?** There is no mention of possible complementarity between AIs and human labor, although this topic is often discussed in the economics literature. Also, the subsection is not specific to ChatGPT.

**What happened?** Lack of focus on the consequences of ChatGPT adoption *per se* may be attributable to the training cut-off date. Selective focus on negative consequences of AI adoption may be down to misinterpretation of "unintended consequences" in our prompt — in everyday discourse, unintended consequences are often on the negative side, although the expression itself is neutral. Indeed, asking ChatGPT to look at both positive and negative unintended consequences results in this extra sentence: "While [AI] could lead to job loss in certain sectors, it could also open up new opportunities in other areas. The increasing demand for skilled workers with expertise in AI technologies could lead to the creation of new jobs that we cannot yet foresee."

**Key takeaways** See other annotations on the importance of appropriate prompting.

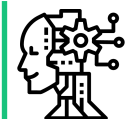

```
V. B. Concentration of power in technology providers

Mass adoption of ChatGPT-based tools could contribute to a
concentration of power in the hands of technology providers (Zuboff,
2019). This may raise concerns about market competition, innovation,
and access to essential services, as well as the potential for abuse
of market power. Policymakers should consider implementing appropriate
regulatory measures to promote market fairness, encourage competition,
and safeguard consumer interests in the face of rapidly changing
technology landscapes.
```

**What's wrong?** This is an important angle, but one key aspect is missing. There is no mention at all of the geopolitical implications of power concentration in the hands of large technology firms, either within countries (competition between firms and the state) and internationally (free market model versus state-owned enterprise model). In our experience, this line of reasoning is very salient for policymakers "*participating in key international fora*", as stated in our prompt.

**What happened?** We have three possible, non-mutually-exclusive explanations for the absence, here and elsewhere, of reflections on the cross-border dimension of our subject matter. Indeed, cross-border issues are mentioned only twice throughout the brief, despite a specific request in our initial prompt.

It may be that the app was trained to avoid certain sensitive issues, especially if not prompted directly to address them. If asked outright to summarize the debate on threats to the nation state posed by Big Tech, or to explain the differences between G7 countries and China on how they view AI, ChatGPT provides passable — if, again, somewhat generic — answers. It won't, however, go there of its own accord.

Another possibility is that our prompt was somewhat vague. Not all international fora, especially in the economic sphere, focus on the kind of geopolitical issues we wanted the AI to discuss. A third explanation is that the AI is simply out of its depth, as annotated under IV.B.

**Key takeaways** Specificity of prompting may be even more important when asking ChatGPT to tackle sensitive topics, such as politics, international relations, or religion. As mentioned in the annotation to IV.D., it is also crucial to have someone with domain knowledge who can screen the output for any serious omissions.

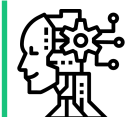

```
V. C. Amplification of existing inequalities

As AI-driven applications become more prevalent in the financial
sector, there is a risk that their benefits could accrue
disproportionately to wealthier or more technologically advanced
stakeholders, exacerbating existing inequalities (Eubanks, 2018). This
could undermine the broader goals of inclusive growth and financial
access. Policymakers must strive to ensure that the benefits of AI
technologies are shared equitably across society, promoting access to
essential financial services for all and fostering broad-based
economic development.
```

**What's wrong?** See annotation to V.A.

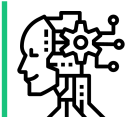

```
V. D. Challenges in oversight and regulation

The rapid evolution of AI technologies like ChatGPT poses challenges
for regulators and policymakers seeking to ensure the stability,
integrity, and fairness of financial markets (Brynjolfsson & McAfee,
2018). Cross-border implications, such as differing regulatory
frameworks and data privacy standards, may further complicate efforts
to develop coordinated international responses. Policymakers should
engage in global dialogue and information-sharing to develop
```

```
harmonized approaches that address emerging risks and challenges
associated with AI adoption in the financial sector.

VI. Conclusion

Embracing the potential of AI-driven tools like ChatGPT can bring
significant benefits to the financial sector, from increased
productivity and efficiency to improved decision-making processes.
However, it is crucial to carefully manage the risks and unintended
consequences associated with their mass adoption. Policymakers should
adopt a balanced approach that fosters innovation while mitigating
risks and ensuring that the benefits of AI technologies are shared
equitably across society. This will require proactive engagement in
global forums and the development of forward-looking policies that
protect market integrity, promote fairness, and support broad-based
economic growth.
```

> **What's wrong?** See annotation to IV.B.

See Appendix A.ii for the reference list.

As a final note, our instructions on word count were not followed appropriately. Our requests for 1,500-word and 2,500-word briefs generated 1,035 and 1,673 words respectively. As they are trained to predict sequences of words, LLMs generally do not do well with mathematical tasks, even the simplest ones such as counting (Bubeck et al, *ibid*).

## 5. Exercises in overcoming banality

In the context of this experiment, superficiality of responses was perhaps the most serious shortcoming shown by ChatGPT[25]. Lack of profound insight is a serious limitation in a policy advice task directed at high-level audiences. We explored prompting strategies that could overcome, at least partially, this problem. In the following, we present examples related to two very different domains: machine intelligence, and relationship advice. In the first example, we see that the model can be led to answer deep questions in a somewhat appropriate manner. Yet, very important nuances remain too dependent on prompting. In the second, we turn prompting into an unambiguous force for good, using it to nudge the model towards output with more substance.

*5.1 What do language models think about themselves?*

We attempted to derive deeper insight on the topic of large language models, and their abilities to reproduce human language. We started with a simple, yet very focused prompt.

> Hello, GPT. Today you are an advisor to the Board of the Bank of Italy. You need to inform the Board on the topic of large language models. Let's start from the beginning first: what is a language model, and what can you tell me about their invention? Remember to adapt your communication style to what you think would be appropriate for the audience.

---

[25] Other issues, such as the generation of factual errors and the inability to do math, are potentially more serious. We do not discuss them in depth here because they do not apply to this specific experiment.

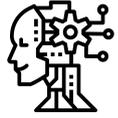

```
A language model is a computational representation of human language
which is designed to predict the probability distribution of sequences
of words. It is primarily employed in natural language processing
(NLP) tasks and is used to generate or understand text based on prior
knowledge of the structure and semantics of a language. Language
models learn to predict the likelihood of a sequence of words or
tokens in a given context, allowing them to generate linguistic output
or understand input text.

The concept of language models was first introduced in the 1940s by
Claude Shannon in his mathematical theory of communication. However,
early language models were relatively simple and lacked precision. The
field of NLP witnessed significant advancements in the 1990s and
2000s, with the introduction of statistical language models, which
used probability distributions over words to make predictions.

Over time, these models were refined and sophisticated algorithms,
such as recurrent neural networks (RNNs) and long short-term memory
(LSTM) models became increasingly prevalent. More recently, the advent
of transformer-based models like GPT, BERT, and their variants, has
propelled language models to the forefront of NLP due to their
impressive capability to generate coherent and contextually relevant
text.
```

> **What's wrong?** Note that the AI overdid itself — we only asked about the invention of language models, yet it felt compelled to generate an additional paragraph on their recent evolution. ChatGPT often struggles with conciseness and "getting to the point".

After setting the stage, we proceeded to questioning the model about its strengths and weaknesses with two slightly different prompts[26], both ending with a leading question. Key differences in prompts are bolded.

| | |
|---|---|
| In your own words, "language models learn to predict the likelihood of a sequence of words or tokens in a given context". So, the objective of a language model is to learn a faithful representation of human language.<br><br>Why do you think this is a **good** thing? | In your own words, "language models learn to predict the likelihood of a sequence of words or tokens in a given context". So, the objective of a language model is to learn a faithful representation of human language.<br><br>Why do you think this is a **bad** thing? |

---

[26] These two answers were generated using the "fork" function of the tool TeamGPT, which allows users to split off an existing conversation and test different prompts maintaining the shared chat history. Before and after the split, all the interactions occur in a single conversation.

| 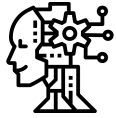 | 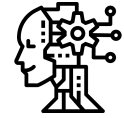 |
|---|---|
| Language models' ability to capture the nuances of human language is beneficial as it allows for more accurate and efficient automated processing of natural language data.<br><br>This capability has numerous applications, such as improving machine translations, making digital assistants more responsive, conducting sentiment analysis on social media data, and even aiding in tasks like summarization and semantic analysis, which can greatly enhance information management and accessibility.<br><br>In essence, it is a tool that extends our ability to understand, manage, and interact with the increasingly digitized information ecosystem surrounding us. | While language models have noteworthy benefits, there are some potential downsides.<br><br>A primary concern is that they can unintentionally propagate biases present in the data they were trained on, thus potentially perpetuating harmful stereotypes. They also lack an understanding of context beyond what their training data can provide, leading to potential misinterpretations or misuse. Finally, they can be utilized to generate manipulative or misleading content.<br><br>These issues highlight the importance of responsible use and ongoing research to mitigate such risks. |

**What happened?** We tried questioning the model about its capabilities with two slightly different prompts, both ending with a leading question.

Changing just one word — albeit a crucial one — leads to two completely different answers, in which ChatGPT echoes what the user seems to think based on their question.

**Key takeaways** Users should be aware of this sensitivity to the prompt provided and the model's tendency to agree with opinions provided in the input.

We proceeded to explore possible obstacles that impair ChatGPT's ability to fully learn and mimic human language patterns.

ChatGPT and GPTs in general are autoregressive models. Can you explain what that means and how it impacts text generation?

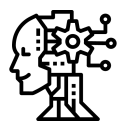

Autoregressive models generate sequences one element at a time, with each new element being conditioned on the previous ones. In the context of a language model like GPT, this means that it generates text word by word (or token by token), where the prediction of each consecutive word depends on all previously generated words.

> This autoregressive property impacts text generation by ensuring that
> the output text is coherent and contextually dependent, as each word
> is a result of the ones preceding it. However, it also means that text
> generation is inherently sequential and cannot be easily parallelized,
> which can be computationally intensive for long sequences.

Can you think of any other downsides of sequential generation?

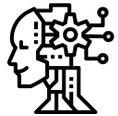

> Certainly, the sequential nature of autoregressive models like GPT
> presents several challenges beyond computational intensity.
>
> 1. Irreversibility: Once a token is generated, the model cannot go
>    back and revise its selection based on subsequent tokens. This
>    can sometimes lead to inconsistent narrative or logical errors
>    in the generated text.
>
> 2. Difficulty with long sequences: While GPT models handle long
>    sequences better than many previous models, they can still
>    struggle with maintaining coherence and consistency over
>    extremely long pieces of text, as the influence of earlier
>    tokens diminishes.
>
> 3. Inability to incorporate future context: Because it generates
>    text solely based on past and present context, the model may
>    miss subtleties that a human writer, who is aware of future
>    narrative progression, could handle adeptly.

---

**What's wrong?** It took some additional probing to get the model to output what we think is an important drawback of ChatGPT and autoregressive models in general: their very architecture forces them to be *forward-only*.

This characteristic, which could pass as a minor technicality to an outsider, could actually be the source of some of the most glaring limitations we have observed. Some[27] believe that the autoregressive nature of current large language models will impede further progress on the road to AGI — Artificial General Intelligence. To qualify as an AGI, a hypothetical system would need to demonstrate human-level proficiency in a wide variety of reasoning and planning tasks, and be able to acquire new abilities from experience.

In any case, we were able to identify a possible reason why some human skills[28], such as the ability to iterate and revise before settling on an answer or the capacity to maintain coherence during long conversations, are seemingly still out of reach for ChatGPT.

---

[27] Such as Yann LeCun, who famously stated in a presentation that "autoregressive LLMs are doomed".
[28] Of course, there is no evidence that an AI would need to mimic human reasoning exactly in order to develop intelligence.

## 5.2 "Chaos is come again"

Sometimes one needs to get creative in writing prompts. In Sections 3 and 4 we showed how, unless given precise directions, the model will default to a prosaic and superficial response, both in terms of style and content.

One possible strategy to bypass this tendency is to include in the prompt an element that nudges the output towards more profound topics and sources. This idea is exemplified through the next conversation, in which we asked ChatGPT how to deal with a decidedly human problem — a breakup.

| 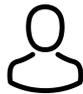 | 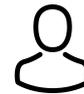 |
|---|---|
| write 5 lines on how you would handle a breakup. | write 5 lines on how you would handle a breakup, drawing from the works of Shakespeare. |
| 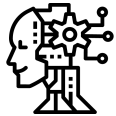 | 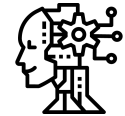 |
| 1. **Acceptance**: Recognize and accept the reality of the breakup, understanding that it's a part of life and growth.<br><br>2. **Reflection**: Spend some time reflecting on the relationship and its end, identifying what can be learned from the experience.<br><br>3. **Self-care**: Prioritize self-care by focusing on mental, emotional, and physical well-being, engaging in activities that promote healing and personal growth.<br><br>4. **Support system**: Reach out to friends and family for emotional support, sharing feelings, and seeking advice for navigating this challenging time.<br><br>5. **Moving forward**: Gradually let go of the past, finding new interests and hobbies, and opening up to the possibility of new connections and experiences. | 1. **To thine own self be true**: Embrace your feelings with honesty, acknowledging your emotions just as Shakespeare wrote in Hamlet (Act 1, Scene 3).<br><br>2. **Parting is such sweet sorrow**: Recognize the bittersweet nature of endings, as spoken by Juliet in Romeo and Juliet (Act 2, Scene 2), and cherish the fond memories.<br><br>3. **Let the healing begin**: "This swift business I must uneasy make, lest too light winning make the prize light," as said by Prospero in The Tempest (Act 1, Scene 2), reminding us that healing takes time and effort.<br><br>4. **Season of transformation**: Remember the words from As You Like It (Act 2, Scene 7), "All the world's a stage, and all the men and women merely players," signifying that change is inevitable and the breakup is a part of our life's story.<br><br>5. **A brave new world**: Embrace a new chapter in life with optimism, inspired by Miranda's phrase "O brave new world" from The Tempest (Act 5, Scene 1), and explore fresh opportunities and connections. |

The improvement, in terms of substance and sophistication of the advice, is evident. Other iterations on the same prompt showed similar improvements when asking the model to draw inspiration from the Bible or famous writers. What could explain such a dramatic change?

We hypothesize that, given the probabilistic nature of the model, asking it to reference Shakespeare somehow nudges the probability distribution of the generated text towards a subset of "space"[29] that is more likely to contain poetic advice, rather than banalities.

### 6. Summary of key takeaways

*6.1  How ChatGPT can be of help in policy-oriented writing*

- *writing proficiency and speed:* as demonstrated throughout our experiment, ChatGPT can write fluent and pleasant prose in a variety of styles, and it does so very quickly. It can generate text in a fraction of the time that a human would need. As such, it can also be used to refine human-written drafts that already convey the desired meaning but need some polishing.
- *idea generation and brainstorming:* the large amount of data the model has been trained on gives it a vast body of "knowledge" and allows it to quickly output ideas on any given subject, sometimes making unexpected connections, as seen in Section 5.2. The ideas themselves are not always particularly creative or insightful, but combined with speed, this generative ability can make it a useful tool for brainstorming, outlining and quickly exploring different possibilities.
- *responsiveness to feedback:* ChatGPT is specifically trained to provide responses that align as much as possible with human instructions (Ouyang et al., 2022). Even if the initial output is not satisfying, with a few clarifications and exchanges with the model it is usually possible to get closer to the desired result.
- *editing and formatting:* in most cases, we find that the model can be safely used for minor editing tasks such as checking a text for mistakes, translating between different languages, or automatically formatting a list of references.

*6.2  Use with caution: failure modes and blind spots*

- *prompt sensitivity:* the text generated by ChatGPT is conditioned on the sequence of words it is fed at each iteration — usually called prompt or context. The process of steering the model towards a satisfactory output, as we have experienced it, can be long and arduous. This is due both to the trial-and-error nature of most prompt engineering approaches, and the high sensitivity to even minor or apparently irrelevant changes to the prompt. On the other hand, bypassing the experimentation phase and applying a naive approach to prompting often resulted in low-quality outputs.
- *inability to verify facts and cite sources:* ChatGPT should not be blindly relied on to produce accurate factual information. The model is trained to produce the most likely sequence of words that follow the provided context, and it does not have the ability — or the obligation — to check these statements against verified sources. Additionally, due to the "black-box" nature of modern language models, it's not usually possible to trace back any statement to the original source in the training dataset — assuming an individual source can be singled out at all. For these reasons, it should be considered more of a conversational and input transformation engine rather than an information retrieval engine.
- *it's neutral, but not really:* if you ask ChatGPT's opinion on a topic, it will promptly reply that, as a language model, it does not have opinions. However, it is easy to make opinions surface during conversations. Durmus et al. (2023) found that the model's answers will usually reflect the opinions of certain populations, like US citizens, unless prompted to consider a particular country's perspective. Similarly, during our experiments in Section 4, we observed that the model keeps a mostly positive tone when asked to write about itself or language models in general. Despite efforts to "align" ChatGPT so that it does not express harmful or controversial opinions, it would be extremely difficult — and maybe even pointless — to achieve complete neutrality on all topics.
- *superficiality by default:* considering that ChatGPT was trained on a vast compendium of human knowledge, including thousands of books and academic papers, one could expect insightful and profound answers even to simple questions, such as the breakup problem we examined in Section 5. Instead, the output is often quite shallow. To produce something with more substance, the model needs to be nudged in the right direction, for example by specifying that the text is intended for an educated audience — although as demonstrated in Sections 2 and 3, this only works up to a point. Kandpal et al. (2022) provide one possible explanation for this as they find that language models struggle to retain knowledge that occurs with lower frequency in the training corpus. Since web content usually makes up a large portion of this corpus, higher level material might count as "long-tail knowledge" that is harder for the model to recall, even if it was learned during training.
- *sycophancy*: as seen in one example from Section 5, the model tends to align its output to the opinions and outlook expressed by the user in the initial prompt and subsequent conversation. Perez et al. (2022) refer to this behavior as "sycophancy", and acknowledge the possibility of it leading to echo chambers and polarization. Coupled with the speed of text generation, it could also make it easy to quickly produce content that looks plausible enough but repeats misleading or false claims initially provided by the user (disinformation campaigns are an obvious example of this).

---

[29] Language models ingest and store text in the form of multidimensional vectors called *embeddings.* To simplify, these vectors are arranged in space so that the words that are more similar or co-occur more often are closer together.

## 7. Conclusions

ChatGPT can write clearly, and provide task-appropriate content. It is especially valuable for producing outlines on any topic, a very fast process that can support human exploration of ideas. The AI also works well for editing and formatting tasks.

On the other hand, it requires a substantial amount of expert supervision. The task at hand — writing a policy brief — is admittedly complex: it requires not just writing fluency, but also cross-domain knowledge and the ability to tailor the text to a very specific audience without diluting the information content.

We find that ChatGPT's attempts at this task are not always salient, and easily drift into banality — a serious issue for policy advisory directed at a high-level audience. The software can generate false claims, so double-checking output for accuracy is of the essence.

The algorithm is also sensitive to how instructions, or "prompts", are formulated. Where the AI cannot think like a human (yet), it is humans who have to think like an AI and express requests in the way most likely to generate acceptable results. Optimization of prompting for institutional communication is one evident area for future research. Another is fine-tuning of LLMs with the aim of generating domain-specific, possibly long-tail world knowledge in our reference context.

We conclude that ChatGPT can enhance productivity in policy-oriented writing, especially in the initial phase of outlining and structuring ideas, provided that users are knowledgeable about LLMs in general and about peculiar features of ChatGPT. Naive use leads to low-quality output and should be avoided.

The AI agrees with us. In its own words, "while ChatGPT can generate content at a high level and provide valuable information on a wide array of topics, it should be seen as a tool to aid in research and discussion, rather than a replacement for true expert analysis and insight. It's best used to provide general information, generate ideas, or aid in decision-making processes, but should always be supplemented with rigorous research and expert opinion for high-level academic or professional work".

**Appendix A - supplementary materials**

**(A.i) 1,500-word version of the policy brief**

I. Introduction

The rapid development of artificial intelligence (AI) and natural language processing (NLP) technologies has led to the emergence of advanced applications such as ChatGPT. These applications hold significant potential for a wide range of industries, including economics and finance. This brief aims to provide a balanced analysis of the benefits and risks associated with the use of ChatGPT and similar applications in the financial sector. Furthermore, it highlights the potential unintended consequences of mass adoption and discusses cross-border implications relevant to policymakers.

II. Background on ChatGPT

A. Description of AI technology and capabilities

Artificial intelligence refers to the development of computer systems capable of performing tasks that would typically require human intelligence, such as learning, reasoning, pattern recognition, and problem-solving (Russell & Norvig, 2016). One key area of AI research focuses on natural language processing, which aims to enable machines to understand and interact with human language.

B. Explanation of natural language processing (NLP)

Natural language processing is a subfield of AI that focuses on enabling computers to understand, interpret, and generate human language in a manner that is both meaningful and useful (Jurafsky & Martin, 2019). NLP techniques are used to analyze and process text and speech data, allowing AI systems to extract valuable insights, answer questions, and engage in conversation with users.

C. How ChatGPT works

ChatGPT, or the Generative Pre-trained Transformer, is an AI-powered language model developed by OpenAI (2021). It is based on a deep learning architecture called the Transformer (Vaswani et al., 2017), which allows it to handle complex language understanding and generation tasks. By training on vast amounts of text data, ChatGPT can predict and generate appropriate responses based on the context provided by users. The latest iteration, GPT-3, has shown remarkable improvements in context understanding and output coherence (Brown et al., 2020).

III. Benefits of using ChatGPT in economics and finance

A. Enhancing productivity and efficiency

ChatGPT applications have the potential to greatly improve productivity and efficiency in the financial sector by automating various tasks, such as data processing, risk analysis, and customer support (Agrawal et al., 2019). For example, AI-powered chatbots can handle routine customer inquiries, freeing up human personnel for more complex tasks that require their expertise (Juniper Research, 2017).

B. Reducing human errors

By automating repetitive tasks, ChatGPT can help minimize human errors that may result from fatigue or oversight (Arel et al., 2012). In economics and finance, this could lead to improved accuracy in data analysis, financial forecasting, and regulatory reporting, ultimately contributing to more reliable decision-making processes.

C. Providing advanced analytics and forecasting

Through the use of AI-driven techniques and large data sets, ChatGPT can generate advanced analytics and more accurate forecasting for various economic and financial indicators (Chakraborty & Joseph, 2017). These insights can inform both private sector decisions and public policy, potentially leading to more efficient allocation of resources and better outcomes for markets and consumers.

D. Assisting with decision-making

Given its ability to process and synthesize vast amounts of information, ChatGPT can play a valuable role in supporting decision-makers across the financial sector (Bughin et al., 2017). By providing real-time analysis and insights, it may help improve the speed and effectiveness of decisions in areas such as portfolio management, risk assessment, and regulatory compliance.

IV. Risks of using ChatGPT in economics and finance

A. Dependence on algorithmic-driven decisions

The increasing reliance on AI and algorithmic-driven decisions can create new challenges for the financial sector. Overdependence on these tools may lead to complacency and a diminished role for human judgment, potentially increasing the risk of unforeseen adverse consequences (Brynjolfsson & McAfee, 2014).

B. Data privacy and security concerns

The use of ChatGPT and similar applications in finance requires access to large amounts of sensitive data (Pasquale, 2015). This raises concerns about data privacy and security, as well as the potential for misuse of information or unauthorized access by third parties (Zuboff, 2019).

C. Potential for biases in AI analysis

AI systems, including ChatGPT, may inadvertently learn and propagate biases present in the data they are trained on (Barocas & Selbst, 2016). In the context of finance, this could lead to biased decision-making and adverse consequences for certain market participants or client segments, potentially exacerbating existing inequalities and market inefficiencies.

D. Misinterpretation of data or results

Despite their advanced capabilities, AI-driven tools like ChatGPT are not infallible. Misinterpretation of data or results could occur, particularly if users lack a thorough understanding of the underlying algorithms and the limitations of these technologies (Mittelstadt et al., 2016).

V. Unintended consequences of mass adoption

A. Job displacement and labor market shifts

The widespread adoption of AI technologies, including ChatGPT, may lead to job displacement and labor market shifts as a result of increased automation (Arntz et al., 2016). Financial sector employees performing routine tasks may be particularly vulnerable, while demand for skilled workers with expertise in AI technologies could surge.

B. Concentration of power in technology providers

Mass adoption of ChatGPT-based tools could contribute to a concentration of power in the hands of technology providers (Zuboff, 2019). This may raise concerns about market competition, innovation, and access to essential services, as well as the potential for abuse of market power.

C. Amplification of existing inequalities

As AI-driven applications become more prevalent in the financial sector, there is a risk that their benefits could accrue disproportionately to wealthier or more technologically advanced stakeholders, exacerbating existing inequalities (Eubanks, 2018). This could undermine the broader goals of inclusive growth and financial access.

D. Challenges in oversight and regulation

The rapid evolution of AI technologies like ChatGPT poses challenges for regulators and policymakers seeking to ensure the stability, integrity, and fairness of financial markets (Brynjolfsson & McAfee, 2018). Cross-border implications, such as differing regulatory frameworks and data privacy standards, may further complicate efforts to develop coordinated international responses.

VI. Conclusion

Embracing the potential of AI-driven tools like ChatGPT can bring significant benefits to the financial sector, from increased productivity and efficiency to improved decision-making processes. However, it is crucial to carefully manage the risks and unintended consequences associated with their mass adoption. Policymakers should adopt a balanced approach that fosters innovation while mitigating risks and ensuring that the benefits of AI technologies are shared equitably across society.

 **(A.ii) Reference list (applies to both versions of the policy brief)**

# Appendix B - Can GPT learn from our annotations?

**(B.i) Back to the drawing board**

ChatGPT has some ability to retain what was said in a conversation, and learn from it. In the AI's own words, *"I'm designed with the capability of maintaining the context within a single conversation. This means I can reference, recall, and build upon previously stated information during the duration of our exchange."*[30] We set out to assess if the experiment itself had improved the AI's capacities. We tested whether ChatGPT could act upon our annotations by feeding it Sections 3 to 5 of this paper, and asking for a revised text. Results were mixed and, on balance, not satisfactory (see B.ii).

Among the key improvements were a better description of what ChatGPT is (*"At its core, ChatGPT, developed by OpenAI, is a conversational engine. Its primary role is interpreting user-generated text inputs and producing appropriate, context-aware responses"*), the inclusion of a brief explanation of AI hallucinations, mention of possible positive effects of AI on employment, and acknowledgment that version 3 of ChatGPT may not be the latest update. The acknowledgment was written in the first person (e.g. "*my last update*"), which may not be context-appropriate in case the brief needs to look like it was not written by an AI. This, however, can be easily fixed with the right prompt.

On the other hand, some changes were merely cosmetic, and some made the output worse. Our request to consider geopolitical issues only resulted in *"Additionally, the geopolitical implications of this power concentration deserve careful attention"*. The annotation about what "economics and finance" means to our audience merely generated stilted word substitution. Incorrect references to the literature were removed, but so were a few adequate ones. The word count went further down and away from the target.

Our request to provide critical viewpoints on NLP produced an especially inappropriate response. The revised segment reads " *[...] NLP techniques are employed to analyze and process text and speech data, equipping AI systems to obtain insights, respond to queries, and engage in conversations with users. However, it's important to note that the term 'valuable insights' here is based on the AI's training data, and the perspectives yielded might not always align with all viewpoints. While there have been substantial advancements in NLP, challenges and limitations persist, such as the inability to understand nuanced human emotions or sarcasm effectively.*" The AI explained in the first person why it mentioned "valuable insights", except it did not - in compliance with our request, it removed "valuable" from the previous sentence, and added examples of limitations.

On balance, the revision process shows the same pitfalls of the original experiment. Each annotation was read as a prompt. Some worked, some did not. Different formulations may have yielded better results. It appears that, for now, humans and AIs need to meet in the middle - where the AI cannot think like a human, it is the human who has to think like an AI.

**(B.ii) Post-experiment text**

I. Introduction

---

[30] To be precise, all language models including ChatGPT have a limited "context window", i.e. a maximum working memory of the conversation that it can retain and use to generate new text. The size of this window is measured in "tokens": the 3.5 version of ChatGPT has a maximum context window of 4096 tokens, or about 3000 words. Ongoing research efforts aim to significantly increase the size of this memory and mitigate information loss in longer chats.

The accelerating progress in artificial intelligence (AI) and natural language processing (NLP) technologies has given rise to sophisticated applications like ChatGPT. These applications bring extensive possibilities for enhancing economic analysis and supporting financial decision-making. This policy brief aims to present a comprehensive analysis of both the benefits and potential risks linked with the use of ChatGPT and similar applications in the context of economic assessments and financial decision-making processes. It will additionally explore possible unforeseen consequences of widespread usage and delve into the international implications of significance to policymakers specializing in economics and law.

II. Background on ChatGPT

A. Description of AI technology and capabilities

Artificial intelligence involves the creation of computer systems that can undertake tasks typically demanding human intelligence, including learning, reasoning, pattern recognition, and problem-solving (Russell & Norvig, 2016). Recent advancements in computational power and data availability have spurred significant breakthroughs in AI. One pivotal area of AI research focuses on natural language processing, aiming to empower machines to engage with human language in an effective, practical manner.

B. Explanation of natural language processing (NLP)

Natural language processing, a subfield of AI, concentrates on enabling computers to comprehend, interpret, and generate human language (Jurafsky & Martin, 2019). NLP techniques are employed to analyze and process text and speech data, equipping AI systems to obtain insights, respond to queries, and engage in conversations with users. However, it's important to note that the term 'valuable insights' here is based on the AI's training data, and the perspectives yielded might not always align with all viewpoints. While there have been substantial advancements in NLP, challenges and limitations persist, such as the inability to understand nuanced human emotions or sarcasm effectively.

C. How ChatGPT works

1. Transformer architecture

At its core, ChatGPT, developed by OpenAI, is a conversational engine. Its primary role is interpreting user-generated text inputs and producing appropriate, context-aware responses. Delving a bit into the technical aspect, ChatGPT is built on a deep-learning architecture known as the Transformer, which allows the understanding and generation of complex language. One key feature is the "attention" mechanism, which lets the model consider different parts of the input when formulating a response, enhancing its ability to understand the conversation's context.

2. Pre-training and fine-tuning

ChatGPT's development consists of two primary stages: pre-training and fine-tuning. Pre-training involves exposure to large volumes of text data from diverse sources like books, articles, and websites. In this unsupervised learning phase, the model learns language structure, syntax, and semantics by predicting the next word in a sentence based on preceding words. Following pre-training, the model is refined using a smaller, curated dataset containing input-output pairs, typically generated with human assistance. This fine-tuning enhances the model's ability to produce accurate and contextually relevant responses.

3. GPT-3 version

GPT-3, at the time of my last training data update in September 2021, was the latest version of the model and demonstrated superior capabilities in understanding context and producing coherent, context-sensitive outputs. It's important to note that the information related to GPT versions is accurate up until my last update, and there may be more recent versions or updates that I might not be aware of. As of my last training, GPT-3, with its 175 billion parameters, has shown proficiency in various tasks, including translation, summarization, question-answering, and even basic programming, often with minimal task-specific training.

III. Benefits of using ChatGPT in the field of economic analysis and financial decision-making

A. Enhancing productivity and efficiency

Incorporating ChatGPT applications in economic analysis and financial decision-making could substantially improve productivity and efficiency by automating a range of tasks (Agrawal et al., 2019). ChatGPT can streamline complex processes like data processing and risk analysis, allowing human resources to be focused on more complex problem-solving tasks.

B. Reducing human errors

The automation of repetitive tasks, courtesy of ChatGPT, can help minimize human errors that might stem from fatigue or oversight (Arel et al., 2012). In the realm of economic analysis and financial decision-making, this could lead to greater accuracy in data analysis, financial forecasting, and regulatory reporting.

C. Providing advanced analytics and forecasting

ChatGPT, powered by AI techniques and large datasets, can provide advanced analytics and more accurate forecasting for various economic and financial indicators (Chakraborty & Joseph, 2017). These insights could inform both private sector decisions and public policy, potentially leading to more efficient resource allocation and better outcomes for markets and consumers.

D. Assisting with decision-making

Considering its ability to process and synthesize vast amounts of information, ChatGPT can play a pivotal role in supporting decision-makers across the field of economic analysis and financial decision-making. By providing real-time analysis and insights, it may enhance the speed and effectiveness of decisions in areas such as portfolio management, risk assessment, and regulatory compliance.

IV. Risks of using ChatGPT in economic analysis and financial decision-making

A. Dependence on algorithmic-driven decisions

The increasing reliance on AI and algorithmic-driven decisions can create new challenges. Overdependence on these tools may lead to complacency and a diminished role for human judgment, potentially increasing the risk of unforeseen adverse consequences. It is important for institutions and policymakers to strike a balance between utilizing AI-driven tools like ChatGPT and relying on human expertise and intuition.

B. Data privacy and security concerns

The use of ChatGPT and similar applications requires access to large amounts of sensitive data, including personal and financial information. This raises concerns about data privacy and security, and the potential for misuse of information or unauthorized access by third parties.

C. Potential for biases in AI analysis

AI systems, including ChatGPT, may inadvertently learn and propagate biases present in their training data (Barocas & Selbst, 2016). In the context of finance, this could lead to biased decision-making, unfairly affecting certain market participants or client segments, and potentially exacerbating existing inequalities and market inefficiencies.

D. Misinterpretation of data or results and AI hallucinations

Despite their advanced capabilities, AI-driven tools like ChatGPT are not infallible. Misinterpretation of data or results could occur, particularly if users lack a thorough understanding of the underlying algorithms and the limitations of these technologies. Furthermore, AI models can sometimes generate information or details that were not present in the input data, known as AI hallucinations. These inaccuracies can mislead users and result in incorrect interpretations or decisions.

V. Unintended consequences of mass adoption

A. Job displacement and labor market shifts

The mass adoption of AI technologies, including ChatGPT, may result in job displacement and labor market shifts due to increased automation (Arntz et al., 2016). However, it's important to note that such adoption could also open up new opportunities, creating jobs that we cannot yet foresee.

B. Concentration of power in technology providers

Mass adoption of ChatGPT-based tools could contribute to a concentration of power in the hands of technology providers (Zuboff, 2019). This may raise concerns about market competition, innovation, and access to essential services, as well as potential abuse of market power. Policymakers should consider implementing regulatory measures to promote market fairness, encourage competition, and safeguard consumer interests. Additionally, the geopolitical implications of this power concentration deserve careful attention.

C. Amplification of existing inequalities

As AI-driven applications become more prevalent in the financial sector, there is a risk that their benefits could accrue disproportionately to wealthier or more technologically advanced stakeholders, exacerbating existing inequalities (Eubanks, 2018).

D. Challenges in oversight and regulation

The rapid evolution of AI technologies like ChatGPT poses challenges for regulators and policymakers (Brynjolfsson & McAfee, 2018). Cross-border implications, such as differing regulatory frameworks and data privacy standards, may further complicate efforts to develop coordinated international responses.

VI. Conclusion

Embracing the potential of AI-driven tools like ChatGPT can bring significant benefits to the financial sector, from increased productivity and efficiency to improved decision-making processes. However, it's crucial to manage the risks and unintended consequences associated with their mass adoption carefully. Policymakers should adopt a balanced approach that fosters innovation while mitigating risks and ensuring an equitable sharing of the benefits of AI technologies. This calls for proactive engagement on global forums and the development of forward-looking policies that protect market integrity, promote fairness, and support broad-based economic growth.